# A MIXED INTEGER LINEAR PROGRAM FOR HUMAN AND MATERIAL RESOURCES OPTIMIZATION IN EMERGENCY DEPARTMENT

**Ibtissem Chouba, Lionel Amodeo, Farouk Yalaoui, Taha Arbaoui**

Institut Charles Delaunay (CNRS FRE 2019), LOSI
Université de Technologie de Troyes,
12 Rue Marie Curie, CS 42060 10004
Troyes cedex, France
ibtissem.chouba @utt.fr, lionel.amodeo@utt.fr,
farouk.yalaoui@utt.fr, taha.arbaoui@utt.fr

**David Laplanche**

Pôle Territorial Santé Publique et Performance ,
Centre Hospitalier de Troyes
101 Avenue Anatole France, 10000, Troyes, France
david.laplanche@ch-troyes.fr

**ABSTRACT:** *The discrepancy between patient demand and the emergency departments (ED) capacity, that mainly depends on human resources and on beds available for patients, often lead to ED's overcrowding and to the increase in waiting time. In this paper, we focus on the optimization of the human (medical and paramedical staff) and material resources (beds) in the ED of the hospital center of Troyes, France (CHT). We seek to minimize the total number of waiting patients from their arrival to their discharge. We propose a mixed integer linear program solved by a sample average approximation (SAA) approach. The program has been tested on a set of real data gathered from the ED information system. Numerical results show that the optimization of human and material resources leads to a decrease of total number of waiting patients.*

## 1 INTRODUCTION

Over the last decades, ED overcrowding has known an important and serious growth around the world. This is due to several hospital's internal and external factors such as: high demand (patient overflow) for services, seasonal epidemics, long waiting time and limited resources. Thus, this phenomenon is known to have adverse impacts on the performance and quality of ED services. To overcome these problems, resources' optimization is an excellent approach by which an efficient system can be achieved [1].

In this context, human and material resource planning of the healthcare system is widely reported in the literature. Several works have introduced simulation, mathematical, linear programming or heuristics models for the desired optimization [1,2,3,4,5,6,7,8,9].

In 2009, Ahmed and Alkhamis [7] proposed a simulation-based optimization approach as a decision support system for the ED in a hospital in Kuwait. The authors aimed to find the optimal staff number that allows to minimize the patients' waiting time. The results provided a decrease of 40% in patients' waiting time. All the levels of the ED process were taken into account in this study. However, optimization model was limited to the staffing of the ED only as other resources were not considered.

In 2012 and 2015, Carbera [1] and Ghanes et al. [6] were interested to minimize the patient's inpatient stay using simulation models. Ghanes et al. [6] proposed a simulation-based optimization model to improve the ED of the French urban hospital Saint Camille taking into account the stochastic aspect of patient's arrival. The new configuration of the ED staff yields a choice of 10% increase in staffing budget resulting 33% reduction of the patient's length of stay (LOS). Also, Carbera [1] proposed an agent simulation approach coupled with an exhaustive search optimization to determine the optimal ED staff configuration (emergency physicians, nurses and administrative staff).

In 2017, Feng et al. [5] presented a new multi-objective simulation optimization algorithm that integrates non-dominated sorting genetic algorithm II (NSGA II) and multi-objective computing budget allocation (MOCBA) in order to determine the optimal medical resources allocation solution that minimize the patient's LOS and the total medical wasted cost.

El-Rifai et al. [2] proposed a stochastic mixed-integer linear program (MILP) solved by a sample average approximation (SAA) approach to allocate the physicians and nurses of the ED of the hospital center of Lille in France while taking into account the stochastic nature of patient's arrivals and service times. The authors aim to minimize the total expected patients' waiting time. Then, to evaluate the resulting personnel schedules, a discrete-event simulation model is used. In this work, only three queues are considered: two assessment queues (patients before and after having auxiliary examinations) operated

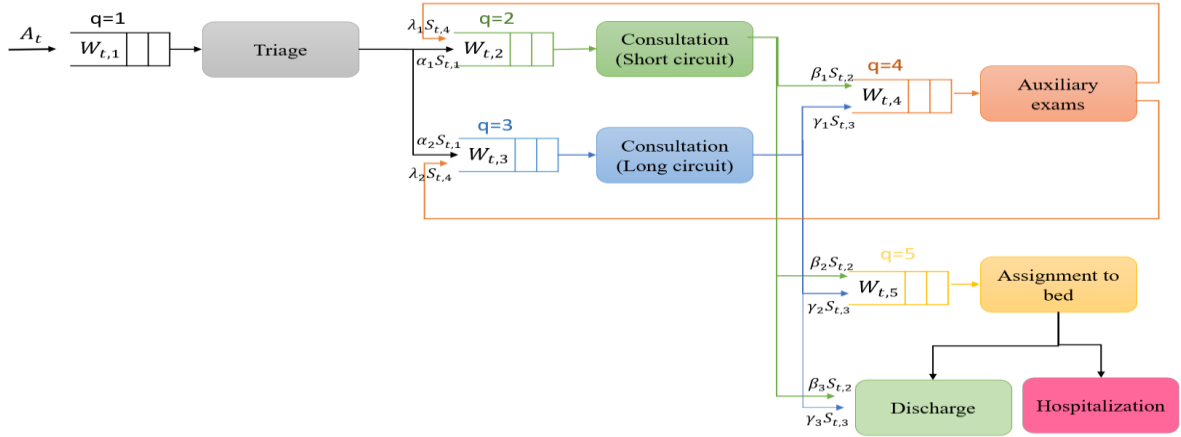

Figure 1 : Patient flow in ED service

by physicians and a third queue operated by nurses for treatment of patients.

In 2018, based on the method proposed by El-Rifai et al. [2], Daldoul et al. [3] used a stochastic MILP solved by SAA to optimize human (physicians and nurses) and also materials (beds) resources organization in the ED at the La-Rabta Tunisian public hospital. They aim to minimize the patient's waiting time, under the uncertainty of the number of patients' arrivals and their service time. In contrast to El-Rifai et al. [2], the authors consider the entire process of an ED from the triage of patient to bed assignment. In the two cases study [2,3], numerical results proved that the patient's waiting time can be decreased by optimizing the number of the human and material resources.

Through the literature review, we note that not all works consider all levels of the process or optimize all the medical and paramedical resources. Also, few studies have utilized the SAA approach to solve the stochastic problems in ED [2,3,8].

In this paper, we focus on the optimization of human (medical and paramedical) and material (beds) resources of the ED of the CHT, France. We aim to increase patients' satisfaction by reducing the number of waiting patients from the triage process to the discharge or the assignment to a bed in the ED. Our model captures all the complexity of the ED process that varies according the patient's state (more details in Section 2). In addition, the uncertainty related to patient's arrival and to the service time are taken into consideration. To achieve this, we develop a stochastic MILP solved by SAA which is one technique of Monte Carlo sampling methods [9]. The proposed model is an extension of the problem proposed by Daldoul et al. [3].

The main contributions of our approach compared to existing work, particularly that of Daldoul et al. [3], are twofold. First, we consider all medical and paramedical resource in the optimization model. Second, to our best knowledge, our work is the first to consider the patient's arrivals forecasts that are introduced as input data in the optimization model. The expected values of patients' arrivals are obtained through forecasting models developed in a previous work [20]. These models proved a very good performance (up to 91,24 % for the annual total flow forecast) and a robustness during epidemic periods [10]. Using these forecasting models, we adapt or relocate necessary resources to meet demand on the long or short term runs.

## 2 MODELLING

Each process in the ED can be modeled as a queue where jobs to be served are the patients and the hospital staff are the servers that handle these jobs [2]. Presented as a queuing system, Figure 1 illustrates the patient's process in the ED of CHT. Two main circuits exist: the short circuit and the long circuit. Within the short circuit, they are mostly autonomous patients, who do not need major paramedical care. Thus, we consider five queues: the first queue (q=1) is for the triage process operated by a triage nurse; the second (q=2) and the third (q=3) queues are for medical consultation in short and long circuit respectively. These queues are operated by the medical (physician, internal) and paramedical (nurse, caregiver) staff. The fourth queue (q=4) is dedicated to auxiliary exams and is operated by nurses. The fifth queue (q=5) is devoted to patients waiting to be assigned to a bed in ED.

In the triage process ($S_{t,1}$), the nurse identifies the patient's state and the pathway they go through: either the short ($\alpha_1$, $S_{t,1}$) or the long ($\alpha_2$, $S_{t,1}$) circuit. Once an appropriate box becomes free, the physician makes a first assessment ($S_{t,2}$ or $S_{t,3}$ for the short and the long circuit respectively) and decides if the patient needs auxiliary exams to confirm or to refine their diagnosis ($\beta_1 S_{t,2}$ and $\gamma_1 S_{t,3}$), to be assigned to a bed ($\beta_2 S_{t,2}$ and $\gamma_2 S_{t,3}$) or to be discharged to go home ($\beta_3 S_{t,2}$ and $\gamma_3 S_{t,3}$). In the case where tests are requested, a second assessment is required in the short or the long circuit ($\lambda_1 S_{t,4}$ and $\lambda_2 S_{t,4}$ respectively).

# 3 MATHEMATICAL FORMULATION

A mathematical formulation based on the details discussed previously is presented in this section. The objective is to increase the patient's satisfaction through the optimization of the number of human and material resources and the minimization of the number of waiting patients.

## 3.1 Parameters

T: Planning horizon
t: Different period of the day [1..T]
I: The number of different types of human resources
i: A type of human resource available [1..I]
Q: The total number of queues
q: The queue number [1..Q]
$N_{i,q}$: The total number of each human resource i, $\forall i \in [1..I], \forall q \in [1..Q], Q \neq 5$
Nb: The total number of beds
n: Number of resources of each type [1..$N_{i,q}$]
$\alpha_j$: Coefficient associated with the patient flow served in the triage box, $\forall j \in [1..4]$
$\beta_j$: Coefficient associated with the patient flow served in the box of consultation in short circuit, $\forall j \in [1..4]$
$\gamma_j$: Coefficient associated with the patient flow served in the box of consultation in long circuit, $\forall j \in [1..2]$
$\lambda_j$: Coefficient associated with the patient flow served in the box of auxiliary exams, $\forall j \in [1..2]$
$TT_i$: The number of total working hours available to schedule resources of type i
LBD: Lower bound of the shift length
UBD: Upper bound of the shift length
$A_t(\omega)$: Random variable representing the number of patients arriving at each period t, under scenario $\omega$.
$L_{i,n,t,q}(\omega)$: Random variable representing the service capacity of the staff, i.e. the number of patients that n resources of type i can serve during period t for each queue q, under scenario $\omega$.

## 3.2 Decision variables

$x_{i,n,t,q} = \begin{cases} 1, \text{ if there are exactly n resources of type i woking at period t.} \\ 0 \text{ else.} \end{cases}$

$p_{i,t,q}$: The number of resources of type i at period t, $\forall q \in [1..Q], Q \neq 5$

$s_{i,t,q}$: The number of resources of type i starting at period t, $\forall i \in [1..I], \forall q \in [1..Q], Q \neq 5$

$e_{i,t,q}$: The number of resources of type i ending at period t, $\forall i \in [1..I], \forall q \in [1..Q], Q \neq 5$

$sb_t$: The number of occupied beds at the start of period t

$eb_t$: The number of liberated beds at the start of period t

$b_t$: The number of beds at period t

$w_{t,q}(\omega)$: The number of waiting patients in queue q at the start of period t, under scenario $\omega$.

$S_{t,q}(\omega)$: The number of patients served in a queue q during period t, under scenario $\omega$.

## 3.3 Resource optimization stochastic model:

$(P): \text{ Min } E_\omega \left[ \sum_{t=1}^{T} \sum_{q=1}^{Q} (W_{t,q}(\omega) - S_{t,q}(\omega)) \right]$ (1)

$W_{1,1}(\omega) = A_{1,q}(\omega)$ (2)

$W_{1,q}(\omega) = 0 \quad \forall q \in [2..Q]$ (3)

$S_{1,q}(\omega) = 0 \quad \forall q \in [1..Q]$ (4)

$W_{t,1}(\omega) = W_{t-1,1}(\omega) + A_t(\omega) - S_{t-1,1}(\omega) \quad \forall t \in [2..T]$ (5)

$W_{t,2}(\omega) = W_{t-1,2}(\omega) + \alpha_1 S_{t-1,1}(\omega) + \lambda_1 S_{t-1,1}(\omega) - S_{t-1,2}(\omega) \quad \forall t \in [2..T]$ (6)

$W_{t,3}(\omega) = W_{t-1,3}(\omega) + \alpha_2 S_{t-1,1}(\omega) + \lambda_2 S_{t-1,1}(\omega) - S_{t-1,3}(\omega) \quad \forall t \in [2..T]$ (7)

$W_{t,4}(\omega) = W_{t-1,4}(\omega) + \beta_1 S_{t-1,2}(\omega) + \gamma_1 S_{t-1,3}(\omega) - S_{t-1,4}(\omega) \quad \forall t \in [2..T]$ (8)

$W_{t,5}(\omega) = W_{t-1,5}(\omega) + \beta_2 S_{t-1,2}(\omega) + \gamma_2 S_{t-1,3}(\omega) - S_{t-1,5}(\omega) \quad \forall t \in [2..T]$ (9)

$S_{t,q}(\omega) \leq W_{t,q}(\omega) \quad \forall t \in [1..T] \quad \forall q \in [1..Q]$ (10)

$S_{t,1}(\omega) \leq \sum_{n=1}^{N_{i,1}} L_{i,n,t,1}(\omega) * x_{i,n,t,1} \quad \forall t \in [1..T] \quad \forall i \in [1..I]$ (11)

$S_{t,2}(\omega) \leq \sum_{n=1}^{N_{i,2}} L_{i,n,t,2}(\omega) * x_{i,n,t,2} \quad \forall t \in [1..T] \quad \forall i \in [1..I]$ (12)

$S_{t,3}(\omega) \leq \sum_{n=1}^{N_{i,3}} L_{i,n,t,3}(\omega) * x_{i,n,t,3} \quad \forall t \in [1..T] \quad \forall i \in [1..I]$ (13)

$S_{t,4}(\omega) \leq \sum_{n=1}^{N_{3,4}} L_{3,n,t,4}(\omega) * x_{3,n,t,4} \quad \forall t \in [1..T]$ (14)

$S_{t,5}(\omega) \leq b_t \quad \forall t \in [1..T]$ (15)

$\sum_{t=1}^{T} p_{i,t,q} \leq TT_i \quad \forall i \in [1..I] \; \forall q \in [1..Q] \; q \neq 5$ (16)

$\sum_{t=1}^{T} (s_{i,t,q} - p_{i,t,q}) = 0 \quad \forall i \in [1..I] \; \forall q \in [1..Q] \; q \neq 5$ (17)

$p_{i,t,q} = \sum_{t'=2}^{t} (s_{i,t',q} - e_{i,t',q}) + p_{i,1,q} \quad \forall t \in [2..T] \; \forall i \in [1..I] \; \forall q \in [1..Q] \; q \neq 5$ (18)

$p_{i,1,q} = p_{i,t,q} + s_{i,1,q} - e_{i,1,q} \quad \forall i \in [1..I] \; \forall q \in [1..Q] \; q \neq 5$ (19)

$\sum_{t'=t+1}^{\text{mod}(t+LBD, T+1)} e_{i,t',q} \leq p_{i,t,q} - s_{i,1,q} \quad \forall t \in [1..T] \; \forall i \in [1..I] \; \forall q \in [1..Q] \; q \neq 5$ (20)

$\sum_{t'=t+1}^{\text{mod}(t+UBD+1, T+1)} e_{i,t',q} \leq p_{i,t,q} \quad \forall t \in [1..T] \; \forall i \in [1..I] \; \forall q \in [1..Q] \; q \neq 5$ (21)

$b_1 = Nb$ (22)

$eb_1 = NB$ (23)

$sb_1 = 0$ (24)

$b_t = b_{t-1} + eb_t - sb_t \quad \forall t \in [2..T]$ (25)

$\sum_{n=1}^{N_{i,q}} n * x_{i,n,t,q} = p_{i,t,q} \quad \forall t \in [1..T] \; \forall i \in [1..I] \; \forall q \in [1..Q] \; q \neq 5$ (26)

$\sum_{n=1}^{N_{i,q}} x_{i,n,t,q} = 1 \quad \forall t \in [1..T] \; \forall i \in [1..I] \; \forall q \in [1..Q] \; q \neq 5$ (27)

$p_{i,t,q} \geq 1 \quad \forall t \in [1..T] \; \forall i \in [1..I]$ (28)

$p_{i,t,q} \geq s_{i,t,q} + 1 \quad \forall t \in [1..T] \; \forall i \in [1..I]$ (29)

The objective function (1) aim to minimize the expected value (E[.]) of the total number of waiting patients while respecting to the distribution of the random value of patient arrival and service times. ω designates one scenario of patient arrival and service time in the ED. Constraint (2), (3), and (4) initialize the number of patients in the queues in each scenario. Constraint (5) to (9) update the number of patients in each queue at every period. Constraint (10) specifies a logical bound for the number of patients served during a period, that cannot be larger than the number of patients in the queue. Constraints (11) through (15) limit the number of patients served in each queue at every period. Constraint (16) limits the total number of staffing hours available for each type of resource. Constraints (17), (18), and (19) are logical constraints that define the range of values that the variables p, s, and e can take. Constraint (20) ensures that at least $s_{i,t,q}$ resources of type i will not end their shift from t to (t + LBD), for each queue q. Moreover, constraint (21) ensures that the last $p_{i,t,q}$ resources of type i end their shift from t + 1 to (t + UBD + 1). Constraints (22), (23), and (24) initialize the number of existing beds, liberated and occupied. Constraint (25) defines the number of available beds at every period. Equations (26) and (27) assign a value to the variable x, which is used to limit the number obtained in constraints (11) through (15). Finally, constraints (28) and (29) are operational constraints to ensure that there are different moments of the day where resources of type i start working and that there is at least one resource of each type in every period.

## 4 NUMERICAL RESULTS

In this section, we present the numerical results of the MILP solved by the SAA method. The main idea of the SAA method is to approximate the expected objective function of the stochastic problem by a sample average estimate derived from a random sample. The resulting sample average approximating problem is then solved by deterministic optimization techniques. The process is repeated with different samples to obtain candidate solutions along with statistical estimates of their optimality gaps [9].

To solve the model, we use the commercial solver IBM ILOG CPLEX 12.5. All the parameters, historical performances and input data are based on real data gathered from the ED's information system and covering the whole year 2019. Based on these data, we use the input analyzer of the ARENA software to generate the distributions of patient's arrivals and services' times in each queue for each period of the scheduling horizon. We assume that the scheduling time horizon is equal 6 hours and the time period (time step) is equal to 1 hour.

To evaluate the performance of the proposed approach, we varied K, the number of scenarios, to obtain K independent random samples of the patient's arrival $(A_t^1…A_t^k)$ and the services' times $(L_t^1…L_t^k)$. The different values of K are [5, 10, 20, 30, 40, 50]. For each of these values, we run the model with seven different replications to produce seven Monte Carlo optimum solutions. The expected total patient's waiting number stabilizes for approximately K = 30 scenarios. We notice that the bigger K is, the better the solution tends towards convergence.

Optimization results are presented in Table 1. We consider 30 scenarios and we present the results of the proposed configurations. First, Table 1 provides the performance (total number of waiting patients) of the current configuration of the ED system which corresponds to the average of the different real scenarios. Second, the performance of the reproduction model (Configuration 1), in which the same number of resources available in the ED are considered, is displayed to evaluate the robustness of our model. Third, the performance of the optimization model (Configuration 2) is shown. Using this model, we measure the impact of the optimizing the number of medical and paramedical resources on each queue $q$. In our experiments, we consider that the maximum number of both medical and paramedical staff per period is equal to 3 (Physician P1=3, Internal P2=3, Nurse P3=3, Caregiver P4=3), and that the number of beds is 17.

Table 1 shows that the total number of waiting patients associated with Configuration 1 (104 patients) is close the real total number of waiting patients (116 patients). For Configuration 2, our model improves the performance by 24% by decreasing the total waiting patients from 104 patients to 79 patients. Optimization results demonstrate that reducing the total number of waiting patients requires adding resources in certain periods and queues.

### 4.1 Incorporating forecasts of patients' arrival

In our work, instead of real arrivals, we introduce patients' arrivals forecasts (Afilal et al. [10]) of the ED of CHT as input data for each period in the optimization model. Following the analysis of patients' arrival data, we noticed that the arrival rate follows a typical behavior whatever the day of the week and the hour of the day. As a result, we can calculate the estimated number of patient arrivals per hour (estimated arrival value* percentage of arrivals per hour). For the resolution, the expected arrival value will be set K times for each period and we generate K independent random samples of the services time.

We test the model, with data of 29-November 2019 (Figure 2) where the mean absolute error between arrival forecasts and observation equal to zero and the number of patients' arrivals was 10% higher than the average daily arrival. This allow as to test the sensitivity and the robustness of the model during an epidemic period for example.

Obtained results are summarized in Table 2. We compute the total number of waiting patients number and the number of required physicians.

Results demonstrate that the total number of waiting patients increases with the increase in the number of patients. Thus, by comparing the optimal configurations, we observe that the number of physicians remains stable in long circuit, except in the first two. However, for the short circuit, we have to add two physicians in average.

|  | **Real simulated data** | | | | | | **Reproduction model (Configuration 1)** | | | | | | **Optimization model (Configuration 2)** | | | | | |
|---|---|---|---|---|---|---|---|---|---|---|---|---|---|---|---|---|---|---|
| Period | T1 | T2 | T3 | T4 | T5 | T6 | T1 | T2 | T3 | T4 | T5 | T6 | T1 | T2 | T3 | T4 | T5 | T6 |
| Triage Nurse | 2 | 2 | 2 | 2 | 2 | 2 | 2 | 2 | 2 | 2 | 2 | 2 | 3 | 3 | 3 | 3 | 3 | 3 |
| Nurse in short circuit | 1 | 1 | 1 | 1 | 1 | 1 | 1 | 1 | 1 | 1 | 1 | 1 | *1* | *1* | *3* | 3 | 3 | 3 |
| Nurse in long circuit | 2 | 2 | 3 | 3 | 3 | 3 | 2 | 2 | 3 | 3 | 3 | 3 | 1 | 1 | 3 | 3 | 3 | 3 |
| Nurse (Auxiliary exams) | 2 | 2 | 2 | 2 | 2 | 2 | 2 | 2 | 2 | 2 | 2 | 2 | 1 | 1 | 1 | 3 | 3 | 3 |
| Triage Caregiver | 1 | 1 | 1 | 1 | 1 | 1 | 1 | 1 | 1 | 1 | 1 | 1 | 1 | 1 | 1 | 1 | 1 | 1 |
| Caregiver in short circuit | 1 | 1 | 1 | 1 | 1 | 1 | 1 | 1 | 1 | 1 | 1 | 1 | 1 | 1 | 3 | 3 | 3 | 3 |
| Caregiver in long circuit | 2 | 2 | 2 | 2 | 2 | 2 | 2 | 2 | 2 | 2 | 2 | 2 | 2 | 2 | 3 | 3 | 3 | 3 |
| Internal in short circuit | 1 | 1 | 1 | 1 | 1 | 1 | 1 | 1 | 1 | 1 | 1 | 1 | 1 | 1 | 1 | 3 | 3 | 3 |
| Internal in long circuit | 2 | 2 | 2 | 2 | 2 | 2 | 2 | 2 | 2 | 2 | 2 | 2 | 2 | 2 | 2 | 3 | 3 | 3 |
| Physician in short circuit | 1 | 1 | 1 | 1 | 1 | 1 | 1 | 1 | 1 | 1 | 1 | 1 | 1 | 1 | 3 | 3 | 3 | 3 |
| Physician in long circuit | 2 | 2 | 2 | 2 | 2 | 2 | 2 | 2 | 2 | 2 | 2 | 2 | 2 | 2 | 2 | 3 | 3 | 3 |
| Beds | 17 | 17 | 17 | 17 | 17 | 17 | 17 | 17 | 17 | 17 | 17 | 17 | 17 | 17 | 17 | 14 | 14 | 13 |
| Performance | 116 patients | | | | | | 104 patients | | | | | | 79 patients | | | | | |

Table 1 : Optimization results for human and material resources

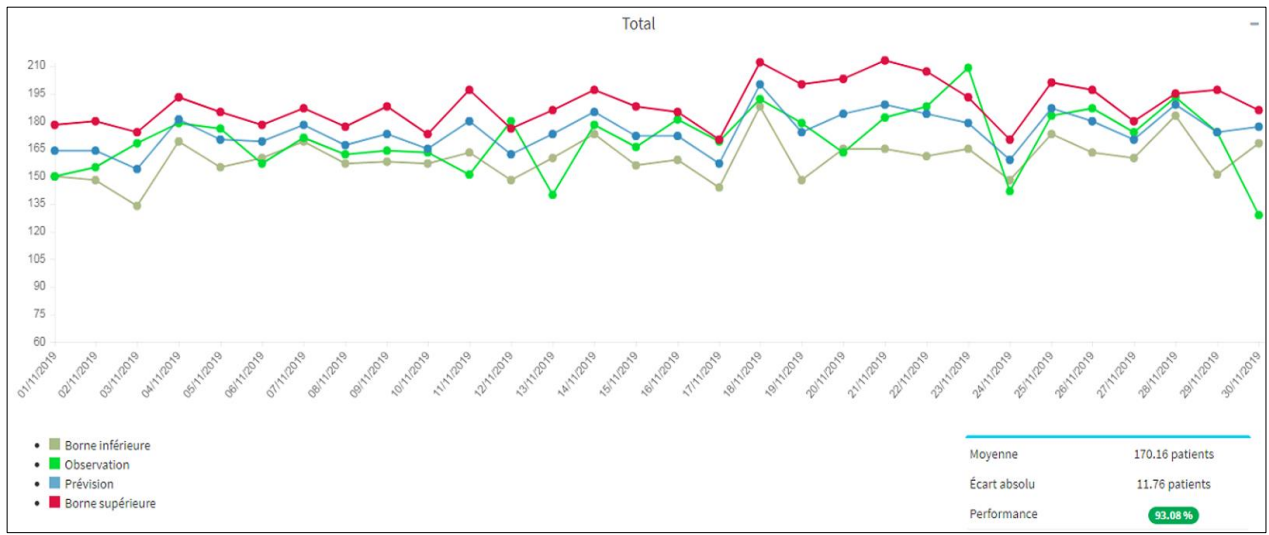

Figure 2: Patient arrival forecasts (Afilal et al [10])

|  | **Physician in short circuit** | | | | | | **Physician in long circuit** | | | | | | **Performance value** | **CPU (seconds)** |
|---|---|---|---|---|---|---|---|---|---|---|---|---|---|---|
| Period | T1 | T2 | T3 | T4 | T5 | T6 | T1 | T2 | T3 | T4 | T5 | T6 | | |
| Current configuration | 1 | 1 | 1 | 1 | 1 | 1 | 2 | 2 | 2 | 2 | 2 | 2 | 146 patients | 1.27 |
| Optimal configuration | 1 | 1 | 3 | 3 | 3 | 3 | 1 | 1 | 2 | 2 | 2 | 2 | 107 patients | |

Table 2: Optimization results for human and material resources considering patient arrival forecasts.

## 5 CONCLUSION

This paper is about the optimization of human and material resources in the ED of CHT. We proposed a generic approach to appropriately adjust the number of human and material resources from the patient arrival into the ED to the discharge or the assignment to emergency beds, taking into account uncertainties related to the patients' arrival and the services times. We proposed a stochastic MILP solved using a SAA method to determine the optimal allocation of medical and paramedical staff and beds, thus minimizing the total number of waiting patients in all the queues. The program was tested on a set of real data gathered from the ED's information system. The experimental results demonstrated that using the proposed approach could reduce the number of waiting patients. In contrast to most existing works, this paper considers optimizing all the humans and materials resources used in the overall patient process. Moreover, to the best of our knowledge, this is the first work that considers the use of forecasts of patients' arrivals as input data to help the decision maker to schedule daily the human and material resources.

In future work, we will consider testing this approach on other emergency structures in order to evaluate their performances and propose new resource organizations and to take into consideration other uncertainties such as the availability of medical staff, the availability of equipment, etc.